\documentclass{llncs}

\usepackage{amsmath, amssymb}

\usepackage{mathtools}

\usepackage{hyperref}
\usepackage{cite}
\usepackage{enumitem}

\usepackage{graphicx}
\graphicspath{{./Figs/}}

\usepackage{siunitx}
\usepackage{booktabs}
\newcommand{\bftab}{\fontseries{b}\selectfont}

\usepackage{multirow}
\usepackage{array}

\usepackage{comment}
\usepackage{caption}
\usepackage{subcaption}
\usepackage[toc,title]{appendix}

\usepackage{bm}
\usepackage{tikz}
\usepackage{dirtytalk}

\newcommand{\ofatdot}{\mathbin{\tikz{\draw[line width=0.25pt] (0,0) circle[radius=0.7ex];\draw[fill] (0,0) circle[radius=0.3ex];}}}

\usepackage{fancyhdr}
\fancypagestyle{firstpage}
{
	\fancyhead{}    
	\fancyfoot[L]{\textbf{The work has been accepted to BMVC 2023.}}
}

\begin{document}
	
\title{Revisiting the Encoding of Satellite Image Time Series}

\author{Xin Cai \and Yaxin Bi \and Peter Nicholl \and Roy Sterritt}
\institute{School of Computing, Ulster University, Belfast, UK\\
	\email{\{cai-x, y.bi, p.nicholl, r.sterritt\}@ulster.ac.uk}}

\maketitle
\thispagestyle{firstpage}
	
\begin{abstract}
Satellite Image Time Series (SITS) representation learning is complex due to high spatiotemporal resolutions, irregular acquisition times, and intricate spatiotemporal interactions. These challenges result in specialized neural network architectures tailored for SITS analysis. The field has witnessed promising results achieved by pioneering researchers, but transferring the latest advances or established paradigms from Computer Vision (CV) to SITS is still highly challenging due to the existing suboptimal representation learning framework. In this paper, we develop a novel perspective of SITS processing as a direct set prediction problem, inspired by the recent trend in adopting query-based transformer decoders to streamline the object detection or image segmentation pipeline. We further propose to decompose the representation learning process of SITS into three explicit steps: collect–update–distribute, which is computationally efficient and suits for irregularly-sampled and asynchronous temporal satellite observations. Facilitated by the unique reformulation, our proposed temporal learning backbone of SITS, initially pre-trained on the resource efficient pixel-set format and then fine-tuned on the downstream dense prediction tasks, has attained new state-of-the-art (SOTA) results on the PASTIS benchmark dataset. Specifically, the clear separation between temporal and spatial components in the semantic/panoptic segmentation pipeline of SITS makes us leverage the latest advances in CV, such as the universal image segmentation architecture, resulting in a noticeable \num{2.5} points increase in mIoU and \num{8.8} points increase in PQ, respectively, compared to the best scores reported so far.
\end{abstract}
	
%-------------------------------------------------------------------------
\section{Introduction}
\label{sec:intro}
Recent years have witnessed a surge of interest in automating the monitoring of the Earth surface based on satellites with high revisit frequency, such as European Space Agency (ESA) Sentinel satellites. In particular, automated large-scale crop type mapping benefits most from leveraging complex temporal dynamics contained in SITS, which can promote the fair allocation of agricultural subsidies and the regulation of the best crop practices being observed by farmers. However, applying deep learning models to extract representative features from SITS is non-trivial, e.g., some of which with a na\"{\i}ve concatenation of spatial and temporal encoders even struggle to surpass the performance of a random forest classifier \cite{kondmann2021denethor}, forcing researchers to devote great efforts to develop bespoke neural architectures. 

The pioneering work PSE+TAE\cite{garnot2020satellite}/PSE+L-TAE\cite{garnot2020lightweight} has introduced a promising learning paradigm for SITS, where statistics of spectral values are first summarized across the spatial extent of crop parcels by Multi-Layer Perceptrons (MLPs) that operate independently on unordered sets of pixels. These summarized spatial features are then fed into a temporal encoder with self-attention to uncover underlying temporal patterns, following a spatio-then-temporal factorization order. With the empirical evidence provided by the recent work TSViT \cite{tarasiou2023vits}, however, it argues that the temporal-then-spatial factorization order is a more intuitive design choice for SITS analysis as spatial contexts in medium-resolution satellite imagery provide non-informative information in contrast to high resolution optical imagery, especially for vegetation monitoring or crop type mapping. This line of research has demonstrated one important aspect when designing deep learning models for SITS: decoupling the learning framework into spatially and temporally separated components. However, the lack of flexibility to operate on different input formats, i.e., the pixel-set or image sequence format, imposes restrictions on PSE+TAE or TSViT. Consequently, the classical pretrain-finetune paradigm in CV, i.e., pre-training a classification model on large-scale datasets (e.g., ImageNet \cite{deng2009imagenet}) with fully-/self-supervised learning \cite{he2022masked, feng2021rethinking} and fine-tuning on downstream tasks such as object detection \cite{ren2015faster} or semantic segmentation \cite{long2015fully}, has not been successfully adopted in SITS analysis yet.

Meanwhile, as pointed out by previous work \cite{garnot2020satellite, garnot2020lightweight}, another great challenge for effectively learning representations for SITS is to capture the complex temporal dynamics in crop phenology, i.e., the precise timings of plant events are crucial for distinguishing various crop types \cite{nyborg2022generalized}. However, recent work for SITS analysis \cite{garnot2020satellite, garnot2020lightweight, nyborg2022generalized, garnot2021panoptic} advocates adopting self-attention \cite{vaswani2017attention} as a core compute unit without questioning its validity for temporal modelling, especially considering its permutation-invariant nature. Based on the latest findings in time series forecasting \cite{zeng2022transformers, woo2022etsformer}, the capability of self-attention operations for modelling complex temporal relations is exaggerated due to a lack of rich semantics in numerical time series data. Modules with strong built-in priors or inductive biases on temporal ordering such as the classical exponential smoothing \cite{woo2022etsformer} or frequency analysis  methods \cite{zhou2022fedformer} have proven to be superior over the vanilla self-attention mechanism for temporal pattern extraction. But irregularity in the temporal axis which is prevalent in satellite image sequences, e.g., optical acquisitions obstructed by clouds, complicates the problem even further, which usually calls for imputation or interpolation as a preprocessing step \cite{kondmann2021denethor} or developing an end-to-end learning framework which should reconcile potentially conflicted optimization objectives \cite{shukla2019interpolation} between interpolation and classification. Except for the validity of self-attention for temporal modelling that has been questioned recently, the quadratic space and time complexity w.r.t. the processed sequence length introduces extra computational concerns for model designs and limits its applicability to dense prediction tasks in SITS \cite{garnot2021panoptic, tarasiou2023vits}. 

These two observations motivated us to reconsider the existing encoding schemes for SITS: \textit{Do we really need to develop bespoke neural architectures for SITS? Is it possible to adapt established CV paradigms to SITS through a simple yet generic representation learning framework?} Specifically, we propose to frame SITS as sets of observations, inspired by the formulation proposed by \cite{horn2020set} for classifying irregularly-sampled and asynchronous time series, where each element is represented by its spectral signatures augmented with static or dynamic covariates such as calendar time or thermal time \cite{nyborg2022generalized}. Facilitated by this unique perspective, we propose a simple yet effective representation learning framework, dubbed as Exchanger, for SITS processing by decomposing the encoding process into three steps: collect--update--distribute, which excludes the use of self-attention to circumvent its limitations. By simply concatenating the proposed Exchanger with a commonly-used segmentation model from CV, we have showcased for the first time that pre-training a classification model on pixel-set format datasets and fine-tuning it on downstream dense prediction tasks with image sequences as input can lead to the new SOTA performance on PASTIS \cite{garnot2021panoptic} compared to highly-specialized network architectures. Furthermore, we can directly introduce the latest universal image segmentation architecture Mask2Former \cite{cheng2022masked} into semantic/panoptic segmentation of SITS without any modifications by simply letting it consume output feature maps encoded by Exchanger, outperforming the previous SOTA models by a significant margin. To sum up, the contributions of this work include:

\begin{itemize}
	
	\item[$\bullet$] redefining SITS representation as sets of instances, eliminating restrictions on model design to accommodate different input data formats of SITS. This allows us to utilize the resource efficient pixel-set format for pre-training, followed by fine-tuning on downstream dense prediction tasks, which we argue is a more desirable way to introduce the pretrain-finetune paradigm from CV to SITS.
	
	\item[$\bullet$] explicitly decomposing the representation learning process of SITS into three steps: collect--update--distribute, leading to a conceptually clear and computationally efficient learning framework, dubbed as Exchanger, for generic feature extraction of SITS.
	
	\item[$\bullet$] in contrast to the existing work where temporal and spatial components are intricately interwoven with each other in the dense prediction pipeline, we argue that a clear separation of temporal and spatial encoders can greatly reduce the complexity in model design and facilitate leveraging the latest advances in CV, mitigating the gap between CV and SITS.
	
	\item[$\bullet$] having conducted extensive experiments to verify the effectiveness of our proposed model, which outperforms the previous SOTA models by a significant margin across semantic and panoptic segmentation tasks on PASTIS benchmark dataset.
	
\end{itemize}
	
%------------------------------------------------------------------------
%------------------------------------------------------------------------
\section{Related Work}
\label{sec:review}	
\textbf{Encoding of SITS} The high frequency revisit time of satellites enables the exploitation of rich temporal dynamics captured for crop type mapping or vegetation monitoring. Traditional machine learning methods \cite{vuolo2018much} rely on handcrafted features where the encoding has not been properly tackled despite the heavy domain expertise required. Recently, differential neural architectures have dominated the field. Specifically, Convolutional Neural Networks (CNNs) \cite{pelletier2019temporal} and Recurrent Neural Networks (RNNs) \cite{russwurm2018multi} have been adopted as a de facto choice to encode spatial and temporal features, respectively. Furthermore, the convolutional-recurrent hybrid models \cite{russwurm2018convolutional} have been proposed to process SITS by viewing it as spatiotemporal signals. Despite the promising results attained, these early attempts have overlooked the significant differences between natural images/videos and SITS. The pioneering work PSE+TAE \cite{garnot2020satellite} has proposed to use MLPs to summarize spatial statistics given the lack of rich spatial semantics in medium-resolution Sentinel-2 images and self-attention to encode temporal patterns, followed by PSE+L-TAE \cite{garnot2020lightweight} where a light-weight transformer decoder has been used to extract temporal features. Pixel-Set Encoder (PSE) is particularly effective for dealing with the irregularity in parcel geometry by simplifying parcel representation from $T\times C\times H\times W$ to $T\times C\times N$, where T is the length of temporal sequence, C is the number channels, H/W denotes the height/width, and N denotes the number of pixels, and consequently requires significantly less storage memory \cite{garnot2020satellite} compared to the patch format. But, when it comes to downstream dense prediction tasks, TAE needs to be integrated into spatial encoders in a complicated manner as shown in the previous SOTA model U-TAE \cite{garnot2021panoptic}, which prevents the replication of the successful pretrain-finetune paradigm. TSViT \cite{tarasiou2023vits} is the first attempt to bridge the gap between SITS analysis and CV by incorporating a unique inductive bias into ViT \cite{dosovitskiy2020image}, which is the temporal-then-spatial factorization based on the observation that spatial contexts provide marginal information for crop type recognition. However, the patch tokenization scheme in ViT is naturally built for images, therefore making TSViT incapable to directly consume unordered pixel-set format, which is a more efficient format for SITS classification and pre-training. Furthermore, the intense computation required by self-attention is exacerbated because the spatial dimension is maintained throughout the whole temporal learning process, which causes TSViT problematic for dense prediction tasks. 
	
%-------------------------------------------------------------------------
\section{Proposed Method}
	
In this section, we first reformulate the representation of SITS as sets of observations in contrast to the conventional spatiotemporal signals. Then, we simplify the current encoding process of SITS by eliminating the need to specially account for the spatial dimension and further decompose the temporal feature learning procedure into three explicit steps: collect--update--distribute. The specific network instantiation is deferred to the supplementary material.

\begin{definition}
	We describe satellite image sequences captured at a particular geo-referenced location with a certain spatial extent as a set $\mathfrak{S}_i$ of instances/sets $\mathfrak{S}_i = \left\{\bm{S}^1, \dots, \bm{S}^n\right\}$, where each instance/set $\bm{S}^j$ is comprised of a set of temporal acquisitions $\bm{S}^j = \left\{\bm{s}_{t_1}^j, \dots, \bm{s}_{t_m}^j\right\}$. And we assume each observation $\bm{s}_{t_k}^j$ is represented by $\left[\bm{v}_{t_k}^j, \bm{p}_{t_k}^j, \ofatdot \right]$, where $\bm{v}_{t_k}^j$ is feature embedding of sensor measurements, $\bm{p}_{t_k}^j$ is temporal positional embedding for a particular acquisition time, and $\ofatdot$ serves as a placeholder for other static or dynamic covariates such as geometric boundaries or modality information, opening up the possibility of arriving at a universal representation for SITS. $\left[ \cdot \right]$ denotes an arbitrary operator to mix the features included in it such as summation or concatenation. Note that the superscript and subscript of $\bm{s}_{t_k}^j$ denote a spatial and temporal identifier, respectively, and we omit the index $i$ for differentiating parcels to avoid notational clutter. 
\end{definition}

In contrast to the commonly-adopted representation of satellite observations as spatiotemporal signals $\bm{\mathcal{X}}_i \in \mathbb{R}^{T\times C\times H\times W}$, we relax the constraints on spatial dimensions imposed by regular grids, for the spatial structure prior is not indispensable for SITS processing \footnote{Note that we restrict the assumption to crop type mapping or vegetation monitoring from SITS. As demonstrated in \cite{jean2019tile2vec}, spatial proximity can be exploited for contrastive representation learning of satellite imagery. Besides, specific land cover recognition, e.g., building footprints, relies most on monotemporal but high resolution imagery \cite{garioud2022flair}.} and further restricts the flexibility when it comes to model design. We argue that more emphasis should be placed on the temporal dimension and the aggregation of spatial information can be flexibly dealt with according to output requirements of various tasks. With such a more universal reformulation, the classification problem of SITS is intimately linked to Multiple Instance Learning (MIL) \cite{ilse2018attention} where a single class label is assigned to a bag of instances with no ordering or strong dependencies among each other, i.e., treating each temporal sequence of observations sampled from different sub-locations within a parcel field as independent instances with uneven contributing weights to the final bag-level classification results. Concerning the dense prediction problem, the regular grid arrangement is only retained for matching the required output format rather than being used for mining high-level spatial semantics. And we have observed in experiments that simply appending well-established semantic segmentation models such as U-Net \cite{ronneberger2015u} after first summarizing temporal information of SITS leads to superior performance to highly-specialized segmentation networks for SITS such as U-TAE \cite{garnot2021panoptic}, which reveals that rich semantics emerge after temporal processing of SITS and resonates with the temporal-then-spatial factorization order advocated in TSViT \cite{tarasiou2023vits}.
	
\subsection{Temporal Context Clusters}
	
\begin{figure}[!htb]
	\centering
	\includegraphics[width=0.85\textwidth]{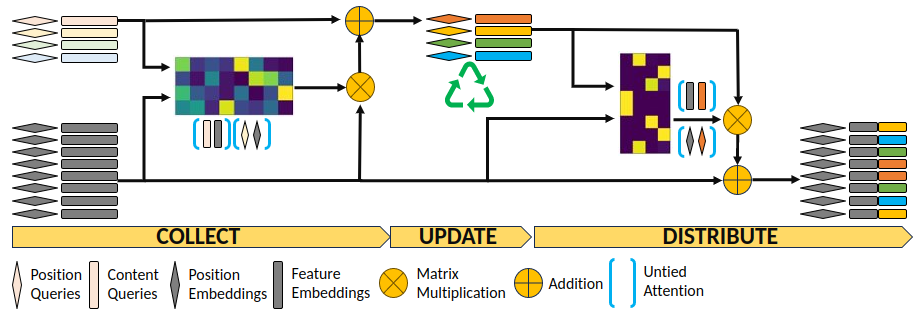}
	\caption{The schematic illustration of the proposed collect--update--distribute procedure for generic representation learning of SITS.}
	\label{fig: 1-1}
\end{figure}

Thanks to our reformulated SITS representation, spatial modeling is not included in the SITS representation learning pipeline due to weak spatial dependencies. As for dense prediction tasks, mining high-level semantics can be accomplished by appending a semantic segmentation model after temporal feature extraction of SITS, which greatly simplifies the existing dense prediction model design for SITS where temporal encoding components are intricately interwoven with spatial encoding components. Motivated by the success of substituting self-attention with other temporal modelling blocks in time series analysis \cite{woo2022etsformer, zhou2022fedformer}, we propose to use a set of learnable queries as an external memory module to exchange temporal information with the input, given that the extra complexity caused by the irregularity in SITS acquisition times, and therefore dub our model \say{Exchanger}. 

Formally, we distil the representation learning process of SITS into three steps: collect--update--distribute, as illustrated in Fig.\ref{fig: 1-1}, with the aid of a set of temporal context clusters, which is further split into two components: content and position queries: $\bm{C}^v \in \mathbb{R}^{N \times d}, \bm{C}^p \in \mathbb{R}^{N \times d}$ to avoid blemishing each other, where $N$ is the number of clusters.

\begin{itemize}
	\item[$\rhd$] \textbf{COLLECT}\hspace{1.5mm} Given the input feature embeddings $\bm{V} \in \mathbb{R}^{T \times d}$ and temporal positional embeddings $\bm{P} \in \mathbb{R}^{T \times d}$, temporal clusters $\bm{C}^{v}$ first collect information from feature embeddings $\left[\bm{v}_1, \dots, \bm{v}_T \right]$ by calculating pair-wise similarities followed by a selective function $\mathcal{S}$ to filter out the least significant ones, which is formulated as follows:
	\begin{align}
		\bm{A}_1 &= \textrm{cal\_simlarity}\left(\left[\bm{C}^v, \bm{V}\right], \left[\bm{C}^p, \bm{P} \right]\right)\nonumber \\
		\bm{W} &= \mathcal{S}\left(\bm{A}_1\right)\nonumber \\
		\bm{C}^v &= \bm{C}^v +  \bm{W}\bm{V} \label{eq:1}
	\end{align}
	where $\bm{A}_1 \in \mathbb{R}^{N \times T}$ is the affinity matrix and is further refined by the selective function $\mathcal{S}$ to obtain $\bm{W}$ to be multiplied by $\bm{V}$, achieving the collection process.
	
	\item[$\rhd$] \textbf{UPDATE}\hspace{1.5mm} Then temporal clusters are updated by solely relying on $\bm{C}^{v}, \bm{C}^{p}$ to allow for global information exchange among different temporal segments, which is formulated as follows:
	\begin{align}
		\bm{C}^v &= \textrm{Update}\left(\bm{C}^v, \bm{C}^p\right) \label{eq:2}
	\end{align}
	
	\item[$\rhd$] \textbf{DISTRIBUTE}\hspace{1.5mm} After updating the clusters, the more robust and representative features of temporal context clusters are distributed back by assigning each temporal element $\bm{v}_i$ to $\bm{C}^v_j$ in a hard or soft manner, which is formulated as follows:
	\begin{align}
		\bm{A}_2 &= \textrm{cal\_simlarity}\left(\left[\bm{V}, \bm{C}^v\right], \left[\bm{P}, \bm{C}^p\right]\right)\nonumber \\
		\bm{I} &= \textrm{assign}\left(\bm{A}_2\right)\nonumber \\
		\bm{V} &= \bm{V} + \bm{I}\bm{C}^v \label{eq:3}
	\end{align}
	where $\bm{A}_2 \in \mathbb{R}^{T \times N}$ is the affinity matrix and each row of $\bm{I} \in \mathbb{R}^{T \times N}$ contains a hard index or soft probability vector to indicate the temporal context cluster to which each temporal element $\bm{v}_i$ is assigned. 
\end{itemize}

The proposed temporal representation learning paradigm collect--update--distribute is particularly effective for dealing with the irregularity and asynchronization in time series data as it imposes no prior assumption such as processing temporal observations in a sequential manner. The features of each temporal element can be updated by interacting with temporal context clusters and information flow among different temporal segments is realized through communication between context clusters, which is a more computationally efficient way for information exchange. Compared to the computation complexity of self-attention $\mathcal{O}\left(T^2d\right)$, it only requires $\mathcal{O}\left(NTd\right)$ where $N \ll T$ and therefore scales much better w.r.t. the number of temporal tokens. More importantly, the proposed representation learning framework for SITS can be seen as a generalization of current self-attention based models such as L-TAE \cite{garnot2020lightweight} or TSViT \cite{tarasiou2023vits}. To be concrete, L-TAE \cite{garnot2020lightweight} is a lightweight transformer decoder where a set of learnable queries is used for extracting key features from outputs of the spatial encoder, which corresponds to the collect step we proposed. The lack of update and distribute steps renders L-TAE less effective for encoding as there is no mechanism implemented for feature updating. The temporal encoder of TSViT \cite{tarasiou2023vits} prepends a set of class tokens to input temporal elements and relies on self-attention for feature learning, which can be seen as a special case of our proposed framework where collect--update--distribute steps are implicitly realized through self-attention. The added external tokens and input temporal elements communicate with each other synchronously, which is more computationally intensive and conceptually vague than our proposed decomposition scheme.
	
\subsection{Network Instantiation}
	
Because of the flexibility of SITS reformulation and the versatility of the proposed collect-update-distribute learning procedure, we chose to draw on recent advances in CV where object queries in the transformer decoder have been reinterpreted as cluster centres and cross-attention has been recast as a clustering operation \cite{suzuki2022clustering, yu2022cmt, cheng2022masked, xu2022groupvit}, reviving the classical idea of framing image segmentation as a pixel grouping procedure rather than per-pixel classification. As clustering is essentially a quantization process where redundant information is gradually filtered out and therefore abstract concepts or high-level semantics may emerge, it has the potential for generic representation learning, not only limited to image segmentation tasks, as demonstrated by the recent pioneering work \cite{yang2022gpvit, ma2023image}. As the main focus of this paper is to establish an effective representation learning framework for SITS, we decided to borrow the core building unit Group Propagation Block (GP Block) from GPViT \cite{yang2022gpvit} to instantiate the idea, leaving the architectural invention for future work. We simply incorporate the construction of GP Block for completeness as follows and refer readers to the original work \cite{yang2022gpvit} for specific details:

\begin{align}
	\bm{C}^v &= \textrm{Concat}_h\left(\textrm{Softmax}\left(\frac{1}{\sqrt{2d}}\bm{C}^v\bm{W}^Q_h\left(\bm{V}\bm{W}^K_h\right)^T + \frac{1}{\sqrt{2d}}\bm{C}^p\bm{U}^Q_h\left(\bm{P}\bm{U}^K_h\right)^T\right)\bm{V}\bm{W}^V_h\right)\label{eq:4}
\end{align}

\noindent where $\bm{W}_h^{Q, K, V}$ and $\bm{U}_h^{Q, K}$ are projection matrices for content and position embeddings, respectively. Eq.\eqref{eq:4} implements the collection process by using cross-attention where the affinity matrix is calculated through scaled dot-product and the softmax function is used for selecting the most relevant temporal elements.

\begin{align}
	\bm{C}^{v} &= \bm{C}^v + \textrm{MLP}_1\left(\textrm{LayerNorm}\left(\bm{C}^v\right)^T\right)^T\nonumber\\
	\bm{C}^v &= \bm{C}^v + \textrm{MLP}_2\left(\textrm{LayerNorm}\left(\bm{C}^v\right)\right) \label{eq:5}
\end{align}

\noindent Eq.\eqref{eq:5} implements the context cluster updating by using a MLPMixer \cite{tolstikhin2021mlp} with one MLPs operated along the token dimension and another MLPs operated along the channel dimension.

\begin{align}
	\bm{Z} &= \textrm{Concat}_h\left(\textrm{Softmax}\left(\frac{1}{\sqrt{2d}}\bm{V}\bm{\tilde{W}}^Q_h\left(\bm{C}^v\bm{\tilde{W}}^K_h\right)^T + \frac{1}{\sqrt{2d}}\bm{P}\bm{\tilde{U}}^Q_h\left(\bm{C}^p\bm{\tilde{U}}^K_h\right)^T\right)\bm{C}^v\bm{\tilde{W}}^V_h\right)\nonumber\\
	\bm{Z}^{\prime} &= \textrm{Concat}\left(\bm{Z}, \bm{V}\right)\bm{\tilde{W}}_{proj}\nonumber\\
	\bm{V}^\prime &= \bm{Z}^{\prime} + \textrm{FFN}\left(\bm{Z}^{\prime}\right) \label{eq:6}
\end{align}
\noindent where $\bm{\tilde{W}}_h^{Q, K, V}$ and $\bm{\tilde{U}}_h^{Q, K}$ are a different set of projection matrices for content and position embeddings, respectively, $\bm{\tilde{W}}_{proj}$ is for linear projection of the concatenated features to the same dimension as the input, and \textrm{FFN} is a feed-forward neural network. Eq.\eqref{eq:6} implements the distribution process by using input temporal elements as queries to gather information from updated context clusters, performing cross-attention in the reversed direction. 

%-------------------------------------------------------------------------
\section{Experiments}
\label{sec: experiments}
In this section, we perform extensive ablation studies to verify the effectiveness of our proposed representation learning framework for SITS and make comparisons with previous SOTA models on semantic and panoptic segmentation tasks. Please note the implementation details are deferred to the supplementary material. The code has been made publicly available at \url{https://github.com/TotalVariation/Exchanger4SITS}.

\subsection{Datasets}
We choose PASTIS (Panoptic Agricultural Satellite TIme Series) \footnote{\url{https://github.com/VSainteuf/pastis-benchmark}} benchmark dataset \cite{garnot2021panoptic} to evaluate the performance of our proposed model and make comparisons with previous SOTA models, which consists of \num[group-separator={,}]{2 433} sequences of multi-spectral images of shape $10 \times 128 \times 128$ and each sequence contains temporal acquisitions taken between September \num{2018} and November \num{2019} with varying sequence lengths between \num{38} and \num{61}, for a total of over \num{2} billion pixels. Furthermore, PASTIS covers four different regions of France with diverse climates and crop distributions, spanning over \SI{4 000}{km^2} and including \num{18} crop types plus a background class. In addition to the spatiotemporal format $T\times C\times H\times W$ with high-quality semantic and panoptic annotations, over \num[group-separator={,}]{120 000} bounding boxes and pixel-precise masks, it is accompanied with a pixel-set format $T\times C\times N$ dataset \cite{garnot2020satellite} for parcel-based crop type classification. We mainly use the 5-Fold splits officially provided by PASTIS for extensive ablation studies and model performance evaluation and additionally report semantic segmentation results on another dataset MTLCC \cite{russwurm2018multi}. The MTLCC dataset covers a large area of interest (AOI) of $\SI{102}{km} \times \SI{42}{km}$ north of Munich, Germany, with \num{17} distinct crop classes and temporal observations of two different lengths of \num{46} and \num{52} gathered in two growing seasons in \num{2016} and \num{2017} \footnote{Please note that the individual samples in MTLCC have limited spatial resolutions of $24\times 24$.}.

\subsection{Implementation Details}

\subsection{Classification} We train and validate the classification model on PASTIS pixelset format dataset. Based on the observation from \cite{wang2022revisiting, sariyildiz2023no} that an additional MLP projector is beneficial for reducing the transferability gap between unsupervised and supervised pre-training, we append the projector proposed in t-ReX \cite{sariyildiz2023no} after the feature extractor Exchanger and use cosine softmax cross-entropy loss. We use AdamW\cite{loshchilov2017decoupled} optimizer, a batch size of $128$, a weight decay of \num{0.005}, an initial learning rate of \num{0.0002}, and a step learning rate scheduler which decays the learning rate at \num{0.7} and \num{0.9} fractions of the total number of training
steps by a factor of \num{10} to train models for \num{50} epochs on \num{4} V100 GPUs. We randomly drop temporal observations by uniformly sampling from the interval between \num{0.2} and \num{0.4} as a data augmentation strategy to counter the adverse effect of cloud obstruction, which has also been adopted in training semantic \& panoptic segmentation models.

\subsection{Semantic \& Panoptic Segmentation} We then use the pre-trained model to initialize Exchanger which serves as the temporal encoder in the semantic/panoptic segmentation pipeline, unless otherwise specified. For the Unet \cite{ronneberger2015u} used as the spatial encoder, we use the AdamW\cite{loshchilov2017decoupled} optimizer, a batch size of $4$, a weight decay of \num{0.005}, an initial learning rate of \num{0.0002}, and a poly decay learning rate scheduler to train models for \num{100} epochs on \num{4} V100 GPUs with Focal cross-entropy loss \cite{lin2017focal} for semantic segmentation and Parcels-as-Points (PaPs) prediction head and PaPs Loss\cite{garnot2021panoptic} for panoptic segmentation. As it cannot fit a single input SITS sample with a spatial resolution of $128\times 128$ and the temporal length of more than \num{30} into V100 GPU with 16G memory, we perform random crop with a crop size of $32\times 32$ in training and test the model performance on full resolution on a A100 GPU. For concatenating the Exchanger with Mask2Former\cite{cheng2022masked} framework, we mainly follow the settings in \cite{cheng2022masked} only with the learning rate changed to \num[scientific-notation = true]{0.00002}. And we train models for \num{100} epochs with a random crop size enlarged to $64\times 64$, a batch size of \num{1} on \num{8} V100 GPUs. Please note when evaluating Exchanger+Mask2Former for panoptic segmentation we split the input into four $64\times 64$ patches and stitch the prediction results together \footnote{We found empirically that the panoptic evaluation metric is particularly sensitive to spatial resolution because of the spatial position encoding extrapolation and patch tokenization layer used in ViT \cite{dosovitskiy2020image, beyer2023flexivit}.}.

\subsection{Ablation Studies}

\begin{table}[!htb]
	\renewcommand{\arraystretch}{1.1}
	\setlength{\tabcolsep}{3.0pt}
	\centering
	\caption{Ablation studies of core design choices in Exchanger on PASTIS validation dataset with 5-Fold cross-validation. The figure in parenthesis denotes the number of content/position queries used.}
	\label{tab:4-3-1}
	\begin{tabular}{cc|c|c|c|c|c}
		\toprule
		& & {Precision\%} & {Recall\%} & {F1 Score\%} & {\#Params(M)} & {FLOPs} \\
		\midrule
		w/o Pos. Queries (4) && 80.0+0.8 & 77.0+1.0 & 78.3+0.9 & 0.50 & \SI{117}{G} \\ \cline{1-2}
		w/ Pos. Queries (4) && 83.5+0.6 & 80.9+0.7 & 82.0+0.5 & 0.52 & \SI{125}{G} \\ \cline{1-2}
		Untied Cont. \& Pos. Attention (4) && 83.6+0.6 & 81.1+0.7 & 82.2+0.5 & 0.52 & \SI{125}{G} \\ \cline{1-2}
		Untied Cont. \& Pos. Attention (8) && 83.9+0.5 & 81.7+1.0 & 82.6+0.7 & 0.52 & \SI{138}{G} \\ \cline{1-2}
		Untied Cont. \& Pos. Attention (16) && 83.4+0.4 & 81.3+0.9 & 82.2+0.6 & 0.52 & \SI{164}{G} \\ \cline{1-2}
		2-Stages (8) && 84.3+0.4 & 82.3+0.4 & 83.1+0.3 & 0.94 & \SI{283}{G} \\ \cline{1-2}
		Temp. Self-Attn (8) && 83.8+0.6 & 81.9+1.0 & 82.6+0.6 & 0.55 & \SI{277}{G} \\ \cline{1-2}
		Temp. \& Spatio. && \multirow{2}{*}{\centering \bftab 84.5+0.6} & \multirow{2}{*}{\centering \bftab 82.7+1.0} & \multirow{2}{*}{\centering \bftab 83.4+0.8} & \multirow{2}{*}{\centering 0.95} & \multirow{2}{*}{\centering \SI{332}{G}}  \\
		Self-Attn (8) && & & & & \\
		\midrule
		\bottomrule
	\end{tabular}
\end{table}

We first study the impact of several key design choices in Exchanger on PASTIS validation dataset compared to a strong baseline model where self-attention is employed to process temporal and spatial features as done in TSViT \cite{tarasiou2023vits}. As seen in Tab.\ref{tab:4-3-1}, not incorporating position queries results in the worst performance with around an absolute $4\%$ drop compared to all other models, indicating date-specific temporal embeddings are key to capture crop phenological profiles. Instead of mixing the content and position information in attention calculation, adopting untied cont. \& pos. attention as proposed in TUPE\cite{ke2020rethinking} slightly improves F1-Score by $0.2\%$, which is set to the default choice for all the subsequent experiments, unless stated otherwise. Then we evaluate the performance of Exchanger w.r.t. the number of content \& position tokens by increasing it from \num{4} to \num{8} to \num{16}. As shown in Tab. \ref{tab:4-3-1}, Exchanger has achieved the best scores across precision, recall and F1 metrics with \num{8} tokens. In contrast to the only \num{1} class token prepended to the input sequence in NLP, we hypothesize that requiring slightly more tokens for crop type recognition is due to the significant intra-class variation and multi-mode nature which we will show the latent embeddings in supplementary materials. Contradicting with fixing the number of tokens to that of classes needed to be identified in TSViT \cite{tarasiou2023vits}, we found that continually increasing the number of content/position queries did not bring the expected performance boost but with a noticeable increase in computational cost. When comparing untied cont. \& pos. attention (8) with its self-attention counterpart (Temp. Self-Attn (8)), it shows that Exchanger can achieve nearly identical results with a similar number of parameters but with a drastic drop in computational cost (almost $50\%$ saving in GFLOPs). Last, with stacking of two identical Exchanger blocks (2-Stages (8)), it reached a F1-Score of \num{83.1}, which is on par with that obtained by Temp. \& Spatio. Self-Attn (8) which is a modified TSViT\cite{tarasiou2023vits} whilst being computationally-light (around $15\%$ saving in GFLOPs). Additionally, the latter (Temp. \& Spatio. Self-Attn (8)) can be seen as adding an attentive MIL pooling component \cite{ilse2018attention} after the temporal self-attention block to identify key spatial instances. However, we have demonstrated solely increasing the depth of Exchanger can bring a similar performance boost, enjoying the advantage that it can be reused in downstream tasks rather than being discarded in TSViT\cite{tarasiou2023vits} for dense prediction. 

\subsection{Convergence Analysis}

\begin{figure}[!htb]
	\centering
	\includegraphics[width=\textwidth]{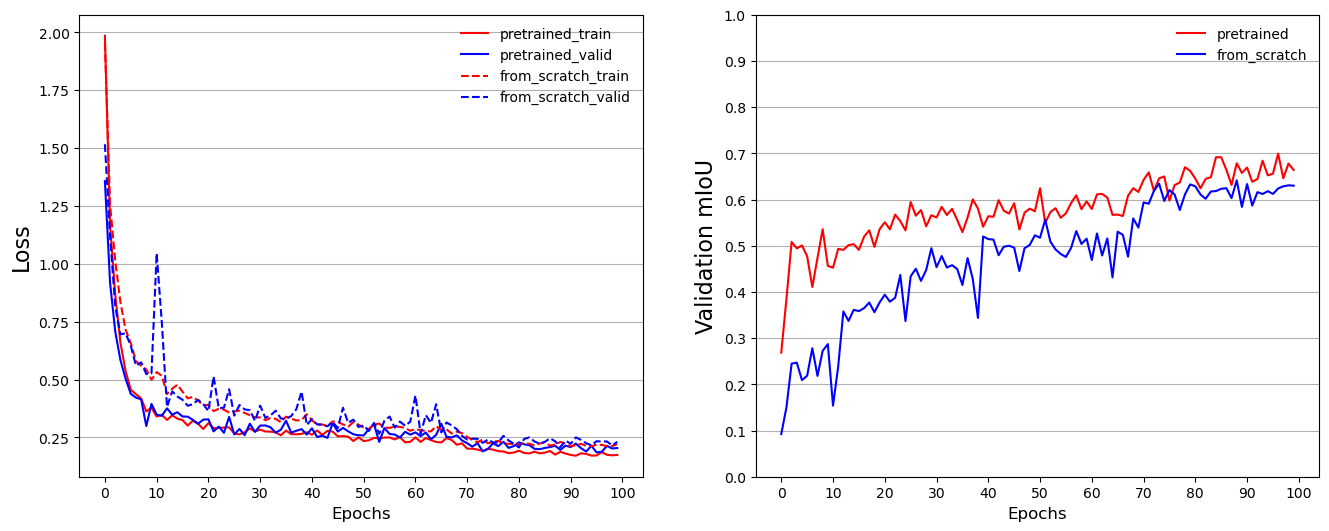}
	\caption{Convergence analysis for Exchanger+Unet with pre-trained backbones or training from scratch on PASTIS validation dataset (Fold-1). The left figure shows the training and validation losses. The right figure shows the evaluation metric mIoU on the validation dataset.}
	\label{fig: 3-1}
\end{figure}

We demonstrate the successful transfer of the pretrain-finetune paradigm from CV to SITS analysis, which is enabled by the reformulated SITS representation, shifting from spatiotemporal signals to sets of instances. It allows the backbone network to be pre-trained on efficient pixel-set format and then fine-tuned on standard spatiotemporal grids for downstream dense prediction tasks. Specifically, as shown in Fig. \ref{fig: 3-1}, pre-trained Exchanger as backbone network appended with a commonly-used segmentation model Unet with randomly initialized weights has led to faster convergence, more stable training and higher validation accuracy than completely training from scratch.

\subsection{Comparison with SOTA}

\subsubsection{Semantic Segmentation}

\begin{table}[!htb]
	\renewcommand{\arraystretch}{1.1}
	\setlength{\tabcolsep}{3.0pt}
	\centering
	\caption{Comparison with SOTA models on PASTIS and MTLCC test dataset. The figure in parenthesis denotes the standard deviation across the official 5-Fold splits in PASTIS. FLOPs are calculated based on a single SITS sample with $T\times C\times H\times W = 30\times 10\times 128\times 128$.}
	\label{tab:4-4-1}
	\begin{tabular}{cc|cccc|c|c}
		\toprule
		&& \multicolumn{4}{c|}{mIoU (\%)}& \multirow{2}{*}{\#Params(M)} & \multirow{2}{*}{FLOPs} \\ \cline{3-6}
		&& \multicolumn{2}{c|}{PASTIS} & \multicolumn{2}{c|}{MTLCC} & & \\
		\midrule
		FPN + ConvLSTM\cite{martinez2021fully} && \multicolumn{2}{c|}{57.1} & \multicolumn{2}{c|}{73.7} & 1.45 & \SI{714}{G} \\
		Unet + ConvLSTM\cite{m2019semantic} && \multicolumn{2}{c|}{57.8} & \multicolumn{2}{c|}{76.2} & 2.33 & \SI{55}{G} \\
		Unet-3D\cite{m2019semantic} && \multicolumn{2}{c|}{58.4} & \multicolumn{2}{c|}{75.2} & 1.55 & {92G} \\
		U-TAE\cite{garnot2021panoptic} && \multicolumn{2}{c|}{63.1} & \multicolumn{2}{c|}{77.1} & 1.09 & \SI{47}{G} \\
		TSViT\cite{tarasiou2023vits} && \multicolumn{2}{c|}{65.4} & \multicolumn{2}{c|}{84.8} & 2.16 & \SI{558}{G} \\
		Exchanger+Unet && \multicolumn{2}{c|}{66.8(+1.2)} & \multicolumn{2}{c|}{\bftab 90.7} & 8.08 & \SI{300}{G} \\
		Exchanger+Mask2Former && \multicolumn{2}{c|}{\bftab 67.9(+1.2)}& \multicolumn{2}{c|}{90.5} & 24.59 & \SI{329}{G} \\
		\midrule
		\bottomrule
	\end{tabular}
\end{table}

As shown in Tab. \ref{tab:4-4-1}, coupling the Exchanger which serves as a pure temporal encoder with a plain Unet \cite{ronneberger2015u} which exclusively focuses on spatial semantic mining has easily led to $66.8\%$ and $90.7\%$ mIoU on PASTIS and MTLCC, surpassing the previous state-of-the-art results attained by TSViT\cite{tarasiou2023vits} by \num{1.4} and \num{5.9} points respectively while only using $53\%$ FLOPs. The dissociation between temporal and spatial components further allows us to explore the potential of adopting the recently proposed powerful universal image segmentation framework Mask2Former\cite{cheng2022masked} with PVT2\cite{wang2021pvtv2} as backbone and FPN\cite{lin2017feature} as the pixel decoder, resulting in a significant improvement of around an absolute $2.5\%$ compared to the best results reported in the literature and a boost of about $1.1\%$ compared to Exchanger+Unet but only with less than $10\%$ increase in the computational cost. It is notable that all previous semantic segmentation models for SITS except for TSViT\cite{tarasiou2023vits} feature a complicated composition of spatial and temporal components, hindering them from leveraging the latest advances in CV. Although TSViT\cite{tarasiou2023vits} is the first fully-attentional neural architecture for SITS processing, it faces extra obstacles when deployed in the pretrain-finetune paradigm because of the patch tokenization layer which prevents it from being directly operated on the pixel-set format, and the self-attention operation can incur prohibitive computational cost for dense prediction tasks. Another marked fact is that the temporal-then-spatial processing order, which has been demonstrated is a more desirable inductive bias\cite{tarasiou2023vits} for SITS analysis, would cause the temporal encoder to consume a drastic proportion of the requested computation, e.g., the Exchanger accounts for nearly $96\%$ of the total computational cost in Exchanger+Unet. And it should be pointed out that our proposed model only has a linear computational complexity $\mathcal{O}\left(NTd\right)$ w.r.t. the input sequence length.

\subsubsection{Panoptic Segmentation}

\begin{table}[!htb]
	\renewcommand{\arraystretch}{1.0}
	\setlength{\tabcolsep}{1.0pt}
	\centering
	\caption{Comparison with state-of-the-art models on PASTIS test dataset. The figure in parenthesis denotes the standard deviation across the official 5-Fold splits in PASTIS. FLOPs are calculated based on a single SITS sample with $T\times C\times H\times W = 30\times 10\times 128\times 128$. Inference Time (IT) is calculated on Fold-1 with around \num{490} sequences on a single A100 GPU.}
	\label{tab:4-4-2}
	\begin{tabular}{cc|c|c|c|c|c|c}
		\toprule
		& & {SQ} & {RQ} & {PQ} & {\#Params(M)} & {FLOPs} & {IT(s)} \\
		\midrule
		Unet+ConvLSTM+PaPs \cite{garnot2021panoptic} && 80.2 & 43.9 & 35.6 & 2.50 & \SI{55}{G} & 660\\
		U-TAE+PaPs \cite{garnot2021panoptic} && 81.5 & 53.2 & 43.8 & 1.26 & \SI{47}{G} & 207\\
		Exchanger+Unet+PaPs && 80.3(+0.1) & 58.9(+0.6) & 47.8(+0.4) & 9.99 & \SI{301}{G} & 252\\
		Exchanger+Mask2Former && {\bftab 84.6(+0.9)} & {\bftab 61.6(+1.6)} & {\bftab 52.6(+1.8)} & 24.63 & \SI{332}{G} & {\bftab 154}\\
		\midrule
		\bottomrule
	\end{tabular}
\end{table}

To further demonstrate the effectiveness of our proposed representation learning framework, we tested its performance on the panoptic segmentation task \cite{kirillov2019panoptic} on PASTIS, which unifies semantic and instance segmentation into a joint task and therefore delivers a holistic scene understanding vision system. Despite the pioneering effort made in \cite{garnot2021panoptic} where a single-stage instance segmentation network CenterMask\cite{wang2020centermask} has been adapted to a panoptic segmentation module named Parcels-as-Points (PaPs), the task still remains extremely difficult as the majority of existing panoptic segmentation networks proposed for natural images or videos is not particularly effective for directly processing SITS. We argue that a strong temporal encoder is key to extracting high-level semantics from SITS, converting the low signal-to-noise ratio 4-D satellite data $T\times C\times H\times W$ to rich semantic 3-D feature maps $C\times H\times W$, which can be fed into off-the-shelf panoptic segmentation models. We report the class-averaged Segmentation Quality (SQ), Recognition Quality (RQ), and Panoptic Quality \footnote{Note that we follow the evaluation protocol in \cite{garnot2021panoptic} where the calculation of PQ only involves thing classes.} (PQ) in Tab.\ref{tab:4-4-2}. It can be seen that Exchanger, equipped with Unet \cite{ronneberger2015u} as the spatial encoder and the PaPs module\cite{garnot2021panoptic} for panoptic prediction, has increased RQ and PQ by a significant margin of $5.7\%$ and $4.0\%$, respectively, compared to U-TAE+PaPs. Furthermore, it is prominent to see that Exchanger combined with Mask2Former\cite{cheng2022masked} consistently outperforms Exchanger+Unet+PaPs by $4.3$, $2.7$ and $4.8$ points in SQ, RQ, and PQ, respectively, setting a new state-of-the-art. Besides, it is noticeable that the required inference time on A100 GPU for Exchanger+Mask2Former is much lower because of the streamlined pipeline and high parallelizability.

\subsection{Qualitative Results}

\begin{figure}[!htb]
	\centering
	\includegraphics[width=0.95\textwidth]{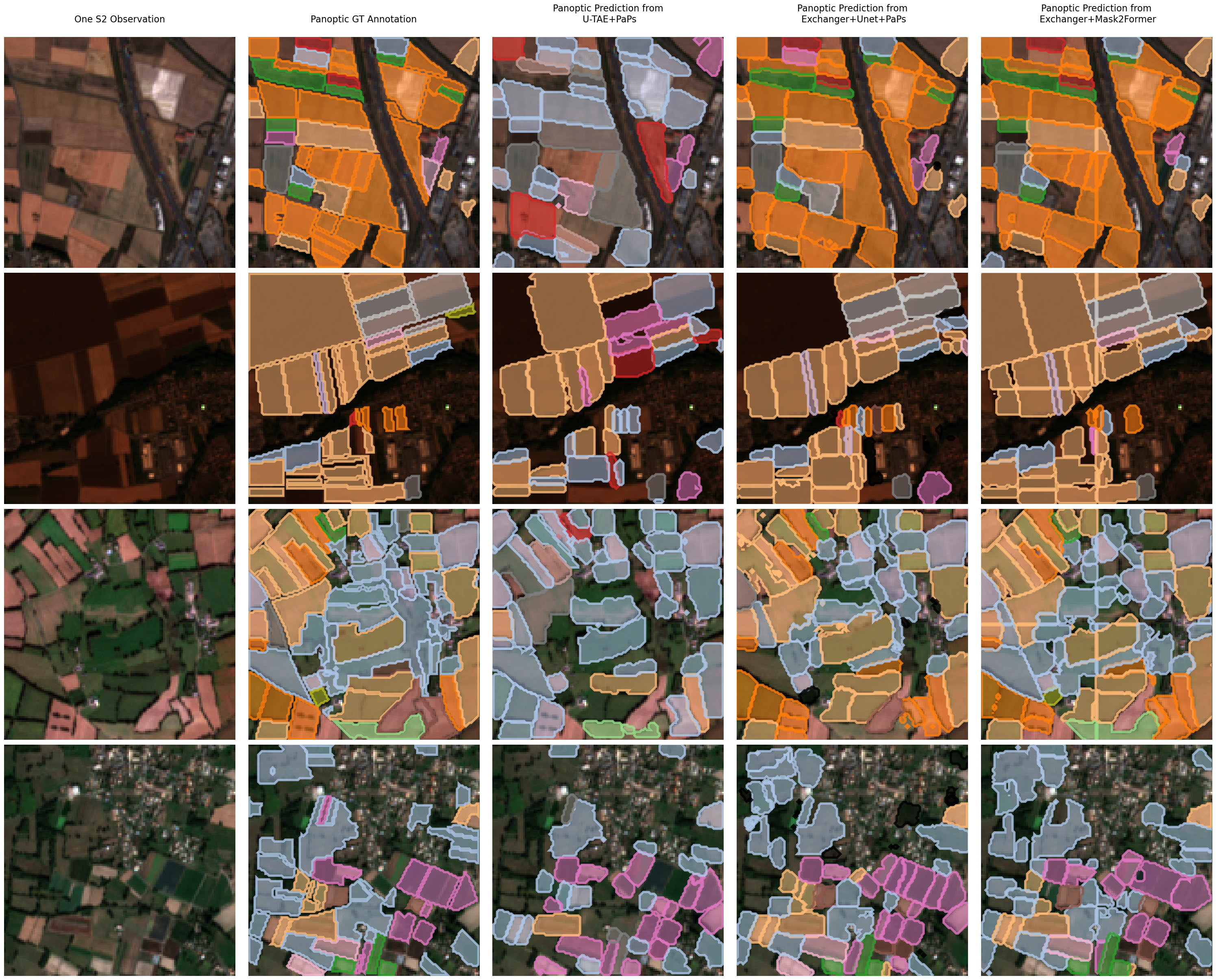}
	\caption{Qualitative comparison. We randomly sample \num{4} SITS sample from PASTIS Fold-1 validation dataset and present the panoptic prediction results from U-TAE+PaPs, Exchanger+Unet+PaPs, and Exchanger+Mask2Former. Please note the artefacts in the last column result from stitching $64\times 64$ predictions to $128\times 128$.}
	\label{fig: 4-5-1}
\end{figure}

In this section, we present a qualitative comparison between previous SOTA model U-TAE+PaPs, Exchanger+Unet+PaPs and the first universal SITS segmentation architecture Exchanger+Mask2Former as a result of concatenating Exchanger as the temporal encoder with the recently proposed universal natural image segmentation framework Mask2Former\cite{cheng2022masked}. As shown in Fig.\ref{fig: 4-5-1}, U-TAE+PaPs can retrieve crop parcels almost as the same number as that of Exchanger+PaPs but is more prone to error predictions, which indicates that the weaker representation learning capability of U-TAE. Coupling Exchanger with a more powerful segmentation architecture Mask2Former\cite{cheng2022masked}, the panoptic prediction quality is significantly improved in terms of crop type recognition accuracy and crop shape prediction consistent with the SQ and RQ metrics reported in Tab.\ref{tab:4-4-2}.

%-------------------------------------------------------------------------
\section{Conclusion}

To conclude, in this paper, we first present a unique reformulation of SITS representation as sets of instances, which relaxes the constraints caused by traditional spatiotemporal grids and further enables designing models that can flexibly process both pixel-set and image sequence format of SITS. Then, we propose to explicitly decompose the representation learning procedure of SITS into three steps: collect--update--distribute, resulting in a conceptually clear and computationally efficient feature learning framework called Exchanger. Facilitated by the previous two innovations, we have demonstrated for the first time the successful transfer of pretrain-finetune paradigm from CV to SITS, leading to a streamlined semantic \& panoptic segmentation pipeline and marked performance gains over the previous SOTA models.  

%%----------------------------------------------------------------
\section*{Acknowledgements}

The work was supported by Department for the Economy (DfE) international studentship at Ulster University (UU). All the experiments presented in the paper were performed at the High Performance Computing (HPC) Centre at UU. We appreciate the constructive and insightful comments of the reviewers.

%%----------------------------------------------------------------
\bibliographystyle{splncs04}	
\bibliography{reference}
\clearpage

%%----------------------------------------------------------------
\begin{appendices}
\appendix
\renewcommand{\thesection}{\Alph{section}.\arabic{section}}
\setcounter{section}{0}
\counterwithin{figure}{section}
\counterwithin{table}{section}

\section{Color Palette for PASTIS}
\begin{figure}[!htb]
	\centering
	\includegraphics[width=0.8\textwidth]{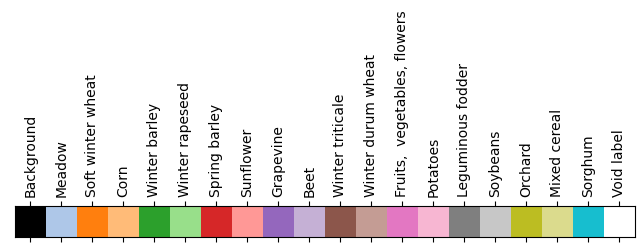}
	\caption{Color Palette used for visualising latent features, semantic \& panoptic predictions on PASTIS.}
	\label{fig: a1-1}
\end{figure}%

\section{Visualisation of the Latent Features in Exchanger}

\begin{figure}[!htb]
	\centering
	\includegraphics[width=\textwidth]{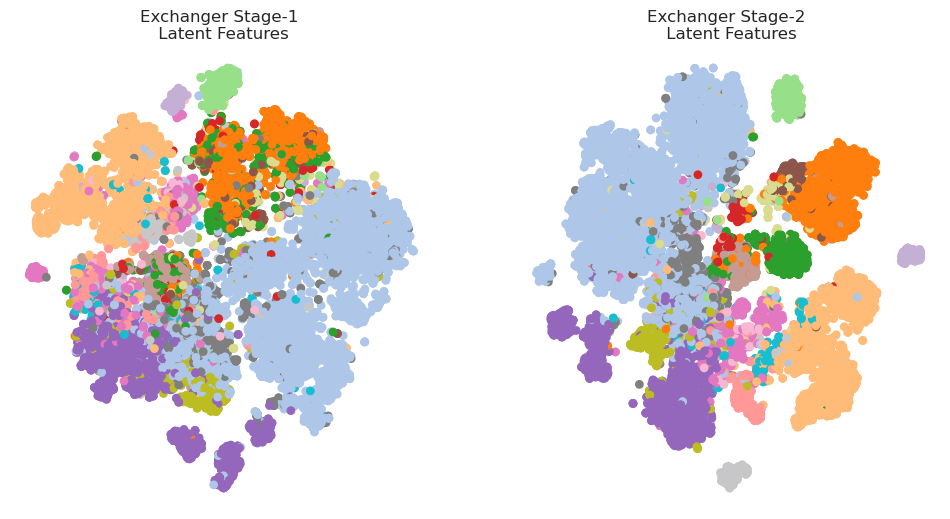}
	\caption{t-SNE \cite{van2008visualizing} visualisations of latent features from stage-1 and stage-2 of Exchanger.}
	\label{fig: b2-1}
\end{figure}

We show latent features from the output of stage-1 and stage-2 of Exchanger before the projector head in Fig.\ref{fig: b2-1}. It can be seen first that the intra-class variation is significantly reduced in the output of stage-2 compared to that of stage-1, indicating a hierarchical clustering procedure enabled by increasing the depth of Exchanger. Additionally, it is noticeable that the multi-mode nature inherited in crop type recognition renders the traditional way in NLP of prepending the input sequence with a single class token ineffective.

\section{More Qualitative Visualisations from Exchanger+Mask2Former}

\begin{figure}[!htb]
	\centering
	\includegraphics[width=\textwidth]{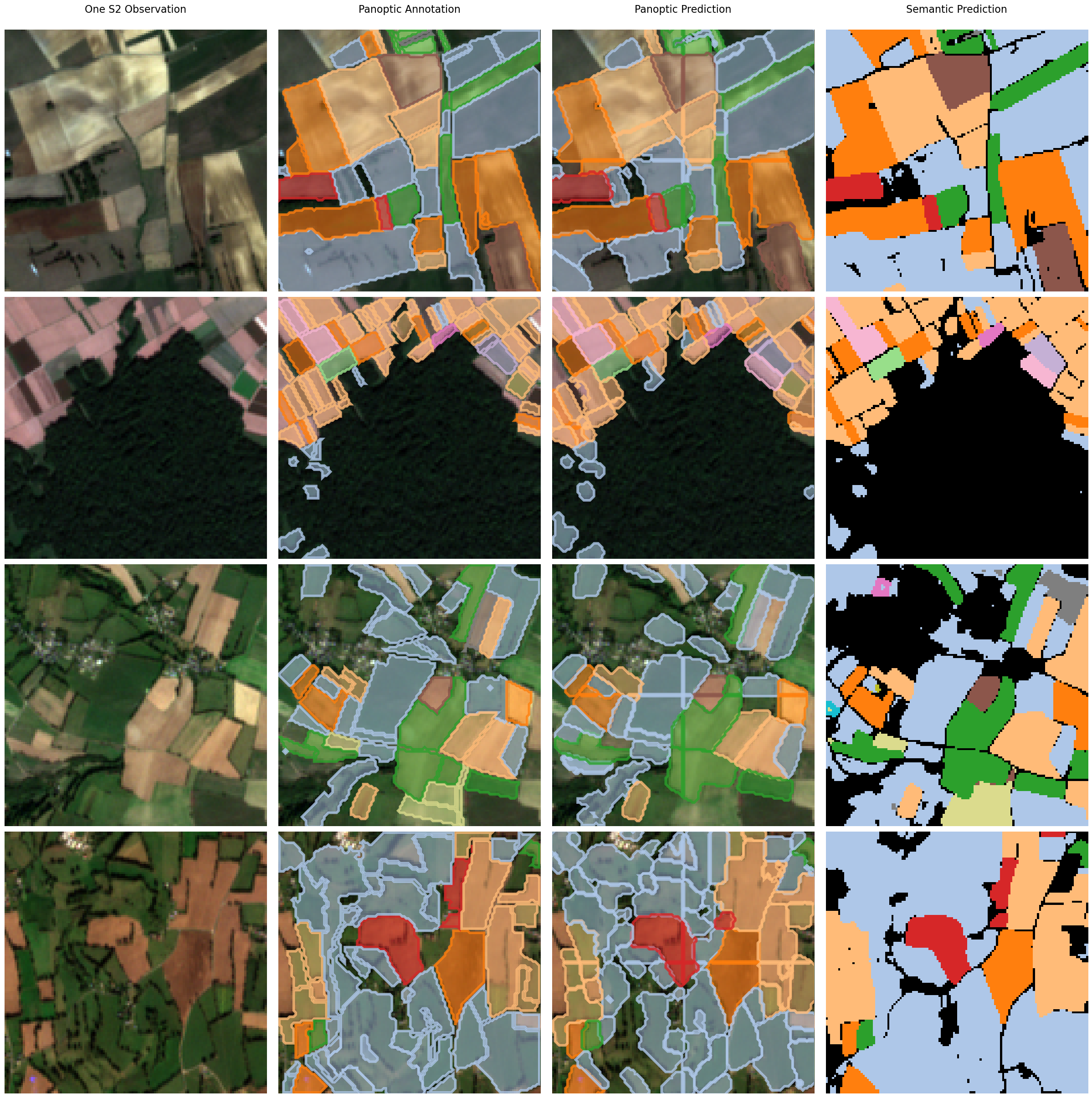}
	\caption{Qualitative Results from predictions of Exchanger+Mask2Former. Please note the segmentation \& panoptic segmentation models are separately trained.}
	\label{fig: c3-1}
\end{figure}

\begin{figure}[!htb]
	\centering
	\includegraphics[width=\textwidth]{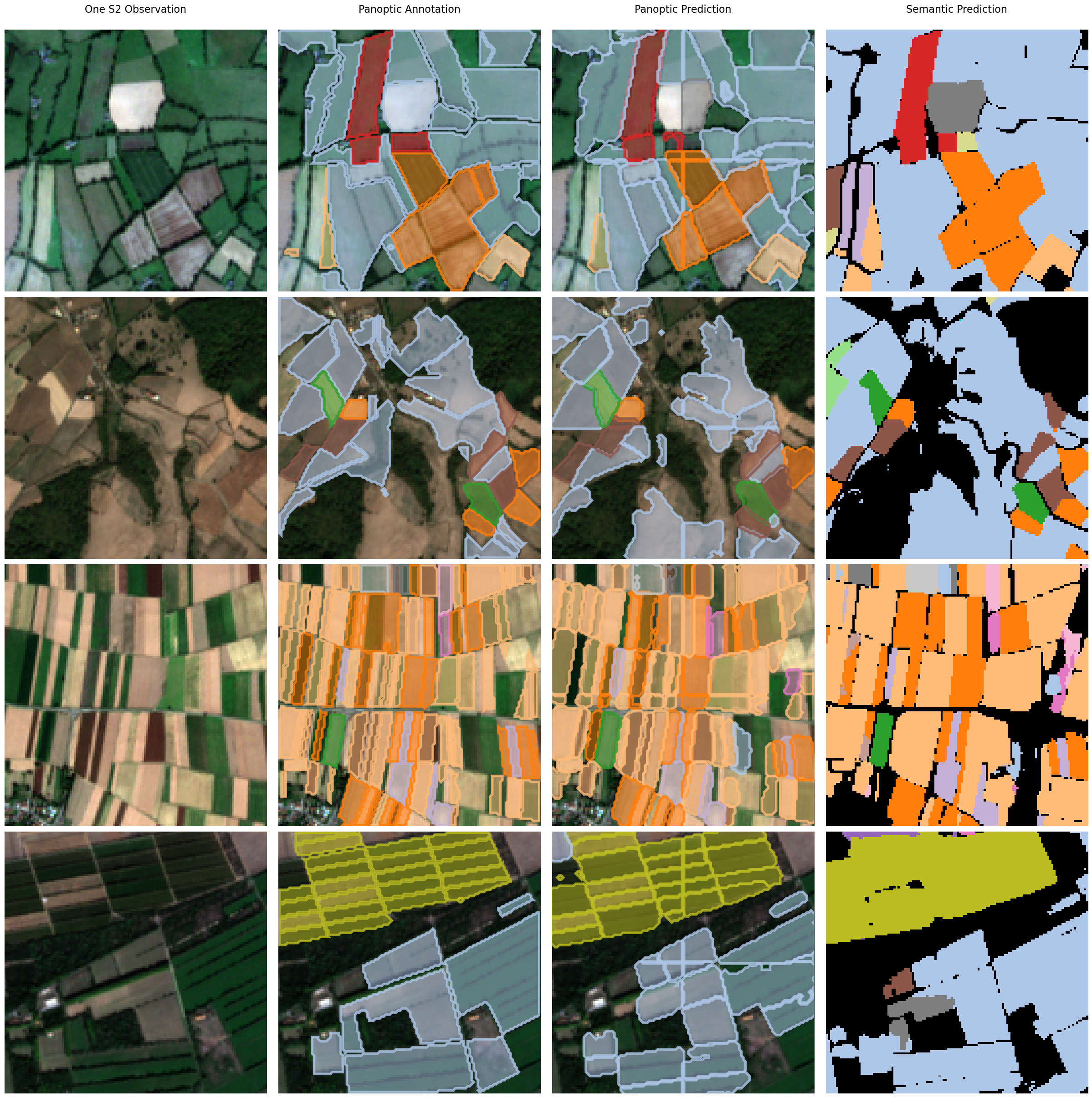}
	\caption{Qualitative Results from predictions of Exchanger+Mask2Former. Please note the segmentation \& panoptic segmentation models are separately trained.}
	\label{fig: c3-2}
\end{figure}

\begin{figure}[!htb]
	\centering
	\includegraphics[width=\textwidth]{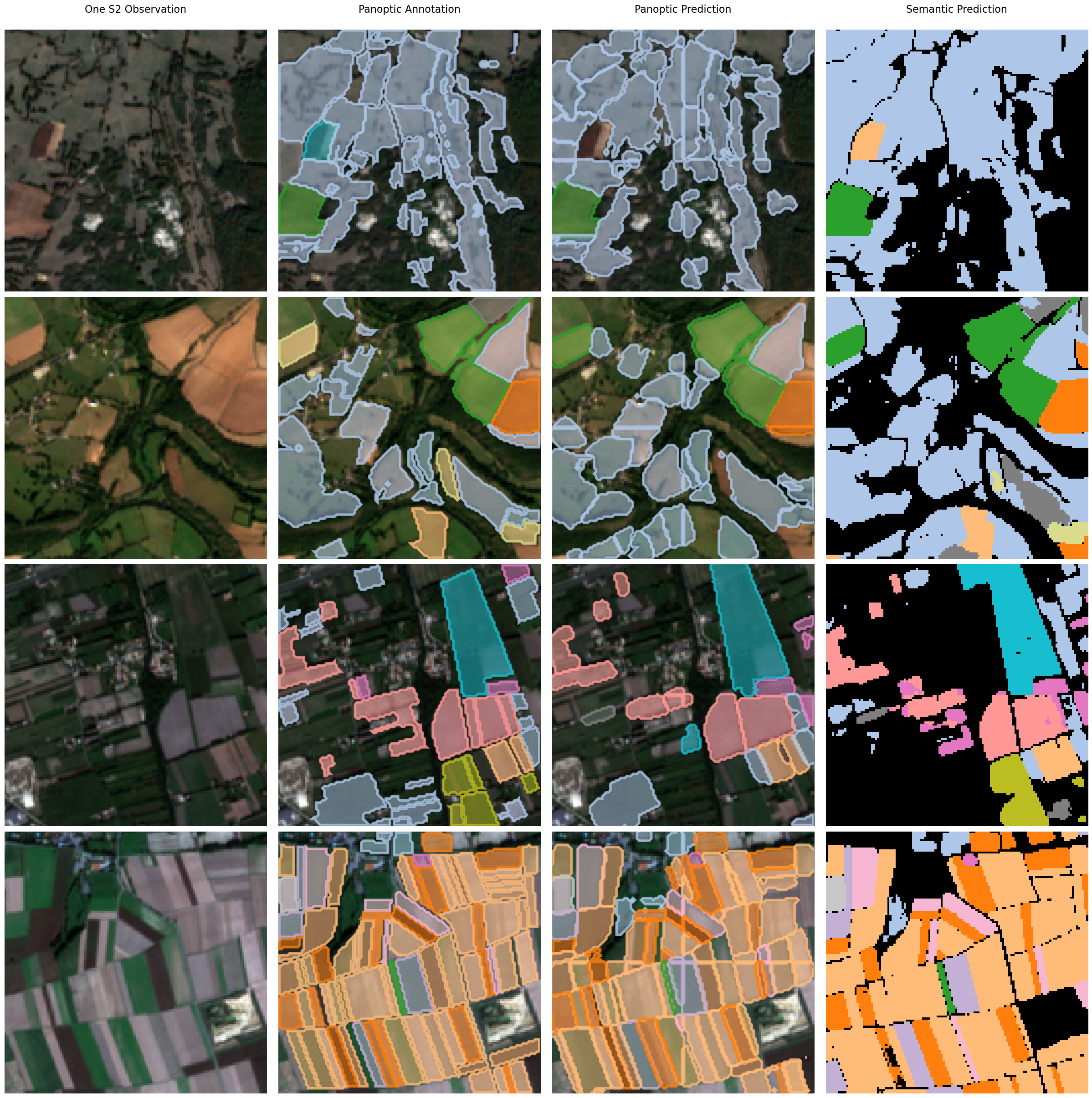}
	\caption{Qualitative Results from predictions of Exchanger+Mask2Former. Please note the segmentation \& panoptic segmentation models are separately trained.}
	\label{fig: c3-3}
\end{figure}

\begin{figure}[!htb]
	\centering
	\includegraphics[width=\textwidth]{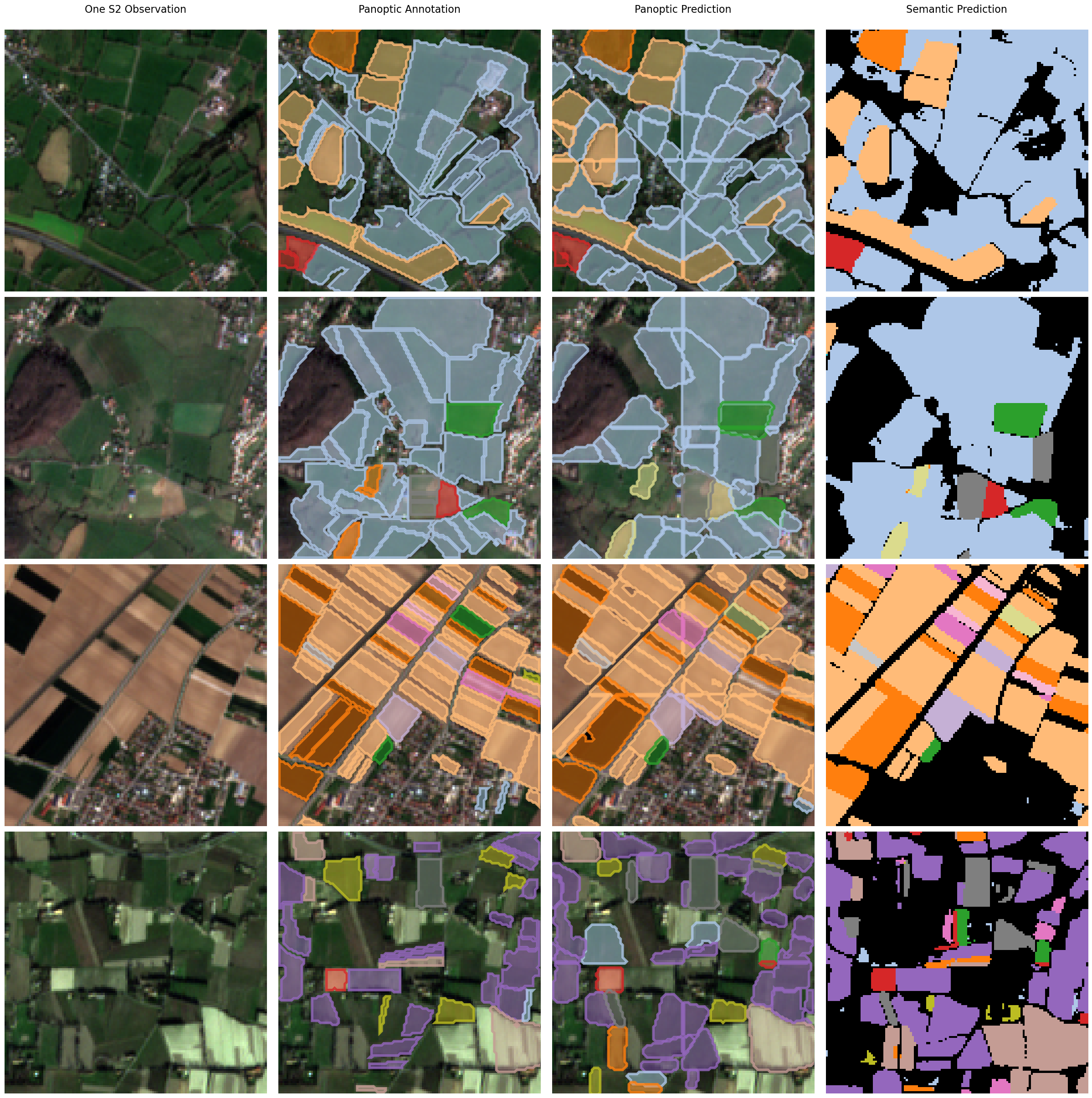}
	\caption{Qualitative Results from predictions of Exchanger+Mask2Former. Please note the segmentation \& panoptic segmentation models are separately trained.}
	\label{fig: c3-4}
\end{figure}
\clearpage

\section{Domain Generalization for SITS}

In this section, we further present results of the Exchanger[2-stages w/ 8 tokens] evaluated on TimeMatch dataset \cite{nyborg2022generalized} which is comprised of SITS from four different tiles: 33UVP (Austria), 32VNH (Denmark), 30TXT (mid-
west France), and 31TCJ (southern France). We follow the naming convention adopted in \cite{nyborg2022generalized} to refer to these four Sentinel-2 tiles as AT1, DK1, FR1, and FR2, respectively, and the leave-one-region-out evaluation protocol where one Sentinel-2 tile is held out for testing and the remaining three tiles are used for training. In addition to the specifically-curated dataset for evaluating spatial generalization capability of crop classifiers, authors in \cite{nyborg2022generalized} proposed to use thermal positional encoding (TPE) to combat temporal shifts across different geographical locations where Growing Degree Days (GDD) have been used to replace calendar time, which has been proven to be effective in improving spatial generalizability. We directly use the TPE method proposed in \cite{nyborg2022generalized} to modify the positional encoding component in Exchanger. Based on our empirical observations, it is favourable to set the dimension of positional embeddings to a relatively small number for better generalization performance, indicating the sensitivity to resolutions of frequencies in sine/cosine functions. As seen in Tab.\ref{tab:d4-1}, our proposed model trained only for \num{20} epochs can achieve results comparable to those of PSE+LTAE\cite{garnot2020lightweight} trained for \num{100} epochs in the original setup. But the highly-specialized architecture PSE+LTAE\cite{garnot2020lightweight} still has demonstrated superiority to our model, which we leave as a future direction for improvement.

\begin{table}[!htb]
	\renewcommand{\arraystretch}{1.1}
	\setlength{\tabcolsep}{3.0pt}
	\centering
	\caption{Leave-one-region-out spatial generalization results (macro F1 score).}
	\label{tab:d4-1}
	\begin{tabular}{ll|c|c|c|c|c}
		\toprule
		&& {AT1} & {DK1} & {FR1} & {FR2} & {Avg.} \\
		\midrule
		\multirow{2}{*}{PSE+LTAE\cite{garnot2020lightweight}}& TPE-Fourier & 84.7 & 79.0 & 77.3 & 80.0 & 80.3\\
		& TPE-Recurrent & \bftab 86.5 & \bftab 80.3 & \bftab 86.0 & \bftab 80.5 & \bftab 83.3 \\\cline{1-2}
		\multirow{2}{*}{Exchanger} & TPE-Fourier & 84.1 & 77.8 & 84.2 & 77.6 & 80.9\\
		& TPE-Recurrent & 82.9 & 80.1 & 81.2 & 76.4 & 80.2\\
		\midrule
		\bottomrule
	\end{tabular}
\end{table}

\end{appendices}

\end{document}